# FlamePINN-1D: Physics-informed neural networks to solve forward and inverse problems of 1D laminar flames

Jiahao Wu, Su Zhang, Yuxin Wu*, Guihua Zhang, Xin Li, Hai Zhang

*Department of Energy and Power Engineering, Key Laboratory for Thermal Science and Power Engineering of Ministry of Education, Tsinghua University, Beijing 100084, China.*

(* Corresponding author. Tel: 010-62799641. Email: wuyx09@tsinghua.edu.cn)

**Abstract:** Given the existence of various forward and inverse problems in combustion studies and applications that necessitate distinct methods for resolution, a framework to solve them in a unified way is critically needed. A promising approach is the integration of machine learning methods with governing equations of combustion systems, which exhibits superior generality and few-shot learning ability compared to purely data-driven methods. In this work, the FlamePINN-1D framework is proposed to solve the forward and inverse problems of 1D laminar flames based on physics-informed neural networks. Three cases with increasing complexity have been tested: Case 1 are freely-propagating premixed (FPP) flames with simplified physical models, while Case 2 and Case 3 are FPP and counterflow premixed (CFP) flames with detailed models, respectively. For forward problems, FlamePINN-1D aims to solve the flame fields and infer the unknown eigenvalues (such as laminar flame speeds) under the constraints of governing equations and boundary conditions. For inverse problems, FlamePINN-1D aims to reconstruct the continuous fields and infer the unknown parameters (such as transport and chemical kinetics parameters) from noisy sparse observations of the flame. Our results strongly validate these capabilities of FlamePINN-1D across various flames and working conditions. Compared to traditional methods, FlamePINN-1D is differentiable and mesh-free, exhibits no discretization errors, and is easier to implement for inverse problems. The inverse problem results also indicate the possibility of optimizing chemical mechanisms from measurements of laboratory 1D flames. Furthermore, some proposed strategies, such as hard constraints and thin-layer normalization, are proven to be essential for the robust learning of FlamePINN-1D. The code for this paper is partially available at https://github.com/CAME-THU/FlamePINN-1D.

**Keywords:** Physics-informed neural network (PINN); Freely-propagating flames; Counterflow flames; Parameter inference

## 1. Introduction

One of the goals of combustion science is to enable human beings to use combustion more efficiently and cleanly [1-3], with the prerequisite that we have a deep understanding of combustion phenomena. However, both combustion studies and applications are challenging because combustion science is an interdisciplinary field that involves nonlinear physical and chemical phenomena across various temporal and spatial scales, including complex chemical reactions and fluid flows [4-6]. Moreover, there exist various forward and inverse problems in combustion studies and applications that require different methods to deal with. In general, forward problems involve applying known combustion theories, models, and technologies to reproduce the combustion process, either virtually or experimentally. This mainly includes combustion simulations for various combustion devices [7-9], the development of numerical methods to solve the governing equations [10, 11], and the application of combustion technologies to achieve controllable combustion [12]. Conversely, inverse problems refer to the development of combustion theories, models, and technologies, which mainly involve theoretical studies based on first principles [6], experimental studies of various combustion phenomena [13, 14], and the identification of the form and parameters of combustion models. Both forward and inverse problems are important, and they are always deeply interrelated. For example, forward numerical simulations cannot be implemented without a complete model, which requires the work of parameter specification, a typical inverse problem. Thus, theoretical, experimental, and numerical methods, which are the three main methodologies for solving these forward and inverse problems, are always adopted together [15]. Considering the intrinsic differences between these three methods that hinder their integration, especially the integration of experimental and numerical methods, the need for a unified framework that can solve both forward and inverse problems is strongly highlighted.

In recent years, artificial intelligence (AI), machine learning (ML), and neural network (NN) methods have attracted considerable attention in the field of combustion due to their success in various applications [16, 17]. These technologies are now considered as the fourth methodology to solve the forward and inverse combustion problems and can also facilitate the integration of theoretical, experimental, and numerical methods. The most commonly adopted paradigm of ML is the data-driven approach, in which ML aims to learn the mapping between labeled variables of interest (supervised learning) or to learn the latent feature of unlabeled data (unsupervised learning), such as clustering and dimensionality reduction. Some examples are briefly listed as follows. (1) For combustion model optimization [18-22], Lapeyre et al. [18] used a convolutional neural network (CNN) inspired by a U-net architecture to predict subgrid-scale flame surface density for a premixed turbulent flame on an under-resolved mesh typical of large eddy simulation. Chung et al. [20] developed a data-assisted modeling approach that uses random forest classifiers to dynamically and locally assign combustion models in reacting flow simulations. (2) For combustion tabulation [23], ML can serve as a promising method to address the memory issue [24-27]. Zhang et al. [24] trained NNs to represent the flamelet generated manifolds (FGM) tables, which significantly reduced the memory requirement. Ranade et al. [25] improved NNs with self-organizing maps (SOMs) to save the memory of multidimensional probability density function tables. The results show that the trained NN requires significantly less memory storage than the lookup table by an order of 1000. (3) For chemistry-related problems, ML can assist mechanism reduction and discovery [28, 29], and predict chemical source terms [30-33], so that the computation of the chemical part in computational fluid dynamics (CFD) simulations can be accelerated. (4) For surrogate modeling of combustion systems, ML is naturally applicable to such tasks since both are a kind of mapping [34-39]. Jung et al. [37] used NNs to predict the combustion characteristics of solid propellants, while Liu et al. [39] used support vector machines (SVMs) and NNs to predict the flame type and liftoff height of fuel jets in turbulent hot coflows.

Despite the noticeable success of data-driven ML for combustion problems, its intrinsic drawbacks of requiring large amounts of data and lack of prior knowledge limit its application to data-few and data-expensive scenarios. To address this issue and to leverage prior knowledge, another paradigm of ML, known as physics-informed ML (PIML), has been proposed [40], where data-driven ML can be integrated with any known priors, such as symmetries and governing equations. PIML can even be conducted without labeled data (zero-shot learning), with the most typical example being the use of NNs to solve ordinary/partial differential equations (ODEs/PDEs) [41]. PIML has been demonstrated to have a stronger ability for extrapolation, interpretation, and few-shot learning than purely data-driven ML. Given that many governing equations of combustion problems are well established, many studies have developed PIML methods in which the ODEs/PDEs are integrated, for both forward (ODE/PDE solving) and inverse problems (such as field reconstruction, parameter inference, surrogate source term). One of the most relevant studies is Neural ODEs [42], where an NN is used to learn the mapping between state variables (temperature $T$ and species mass fractions $Y_k$) and their temporal derivatives (reaction source terms). Based on the ODEs, the NN outputs are numerically integrated to the predicted state variables, which are then optimized to match the true values [43]. The trained NN can subsequently be used in CFD simulation to accelerate ODE solving. Neural ODEs are typically trained in chemical kinetics problems (0D combustion), where the stiffness difficulty has also been addressed [44]. Additionally, Su et al. [45] used the Neural ODE to infer the kinetic parameters from noisy data (a typical inverse problem) and demonstrated the accuracy and efficiency of their framework.

Another class of ML methods integrated with ODEs/PDEs is the physics-informed neural network (PINN), which can solve both forward and inverse problems [41]. In PINNs, an NN maps the spatiotemporal coordinates to the dependent variables. Auto-differentiation (AD) is used to compute the residuals of the ODEs/PDEs and initial/boundary conditions (IC/BCs), which are then substituted into the loss function to be minimized. Compared with traditional methods such as the finite difference method (FDM), PINNs are mesh-free and can seamlessly utilize observed data, so they have attracted much attention and have been widely used in fluid dynamics to solve the Navier-Stokes flow problems [46-50] and multi-species flow problems [51]. The application of PINNs to combustion problems has also been explored, and some studies can be summarized as follows. (1) For 0D problems [52-54], the stiffness of chemical kinetics is the major concern. Ji et al. [54] proposed Stiff-PINN, which is embedded with the quasi-steady-state assumption (QSSA) to reduce the stiffness of the ODE systems. Two classical stiff chemical kinetic problems, ROBER and POLLU, were successfully solved. Wang et al. [52] designed hard constraints on IC for PINN to solve the ROBER problem, while De Florio et al. [53] applied the extreme theory of functional connections (X-TFC) to solve four stiff chemical kinetic problems. (2) For 1D problems [55-57], Liu et al. [57] used PINNs to forward solve the 1D steady premixed flames with detailed combustion models, and the accuracy of PINNs was validated. Hosseini et al. [56] used PINNs to forward solve the Bratu equation arising from solid biofuel combustion theory, while Wang et al. [55] applied PINNs to reconstruct the temperature, velocity, pressure, and chemical reaction progress fields of a rotating detonation combustor (RDC) based on the sparse observation of the first three fields, which is a 1D transient inverse problem. (3) For 2D problems [52, 57-61], Wang et al. [52] used PINNs to solve the 2D mixture fraction equation, while Sitte et al. [61] applied PINNs to reconstruct the velocity fields in puffing pool fires. Chemical reactions are not included in their PINNs. Wang et al. [58] developed a physics-informed generative adversarial network (PIGAN) for the super-resolution flow field reconstruction in RDCs. Liu et al. [57] proposed a parameterized PINN framework for the field-resolving surrogate modeling of 2D laminar premixed combustion, illustrating the advantage of PINN for combustion surrogate modeling over traditional methods. (4) For 3D problems, Choi et al. [62] used PINNs for the transient field reconstruction of a continuous stirred tank reactor with van de Vusse reaction, while Liu et al. [63] used PINNs for the field reconstruction of turbulent flames (freely-propagating premixed flames and slot-jet premixed flames) from sparse noisy data. Both of these applications are inverse problems.

The above studies on solving combustion problems with PINNs still have some limitations and gaps. Firstly, there is no unified framework based on PINN for solving the forward and inverse problems, which is one of the advantages of PINNs. Secondly, 1D laminar flames, which are essential for the study of flame limit [64-66], stretch, laminar flame speed ($s_L$) [67-69], and flamelet tabulation [23], often have unknown eigenvalues [4], such as $s_L$ for freely-propagating premixed (FPP) flames. This makes the problems ill-posed and necessitates additional strategies for traditional methods. However, the 1D laminar flame problems solved by PINNs in previous studies are all well-posed. Specifically, the $s_L$ is explicitly given for FPP flames, which limits the application of PINNs to practical computations. Thirdly, none of the studies involve inverse problems for 1D combustion. It remains open for the feasibility of PINNs to reconstruct the continuous flame fields and learn unknown parameters (such as transport and chemical kinetics parameters) from the noisy sparse observations of laboratory 1D flames.

Therefore, in this work, for the purpose of exploring the potential of PINNs to solve 1D laminar flames with eigenvalues in forward problems, and to reconstruct the flame fields and infer the unknown parameters from noisy sparse observations in inverse problems, the FlamePINN-1D framework has been proposed and applied to various 1D laminar flames, including FPP and counterflow premixed (CFP) flames. The FlamePINN-1D framework seamlessly integrates the governing equations and BCs of 1D laminar flames into the training of NNs, which represent the solution of the corresponding ODE systems. We highlight that FlamePINN-1D is a unified framework that can solve both forward and inverse problems. For forward problems, FlamePINN-1D can solve the flame fields and eigenvalues without any label data, performing the same function as traditional methods. For inverse problems, based on noisy sparse observations of the 1D fields, FlamePINN-1D can reconstruct the continuous fields and infer the unknown transport and chemical kinetics parameters under the constraint of governing equations. To address the challenges posed by strong nonlinearities and multi-physics processes, some strategies are proposed, such as warmup pretraining, hard constraint, thin-layer normalization, and modified NN architecture, which will be discussed in detail in the following sections.

We first present the methodology in Section 2, including the models, the case setup, and the FlamePINN-1D framework. Then, the results are presented and discussed in Section 3, which is divided into the forward and inverse problem parts. Hyperparameter validation and computational efficiency are also discussed. The paper ends with the conclusions in Section 4.

## 2. Methodology

In this section, we first introduce the physical models and configurations of the test cases in this work, explaining how the reference solutions are obtained. Then, the details of the FlamePINN-1D framework for solving the forward and inverse problems of 1D laminar flames are presented, including the basic principles and the transformation of the variables. Finally, the PINN configurations, parameters, and implementation details in this work are provided.

Three cases of increasing complexity were tested: Case 1 describes FPP flames with simplified physical models, while Case 2 and Case 3 are FPP and CFP flames with detailed physical models, respectively. The schematics of the three cases are shown in Fig. 1. Case 1 can be seen as a toy case in which we aim to demonstrate the feasibility of FlamePINN-1D in solving 1D flame problems with Arrhenius reaction and unknown eigenvalues. In Case 2 and Case 3, we aim to apply FlamePINN-1D to problems with more complex combustion models to further demonstrate its comprehensive capabilities.

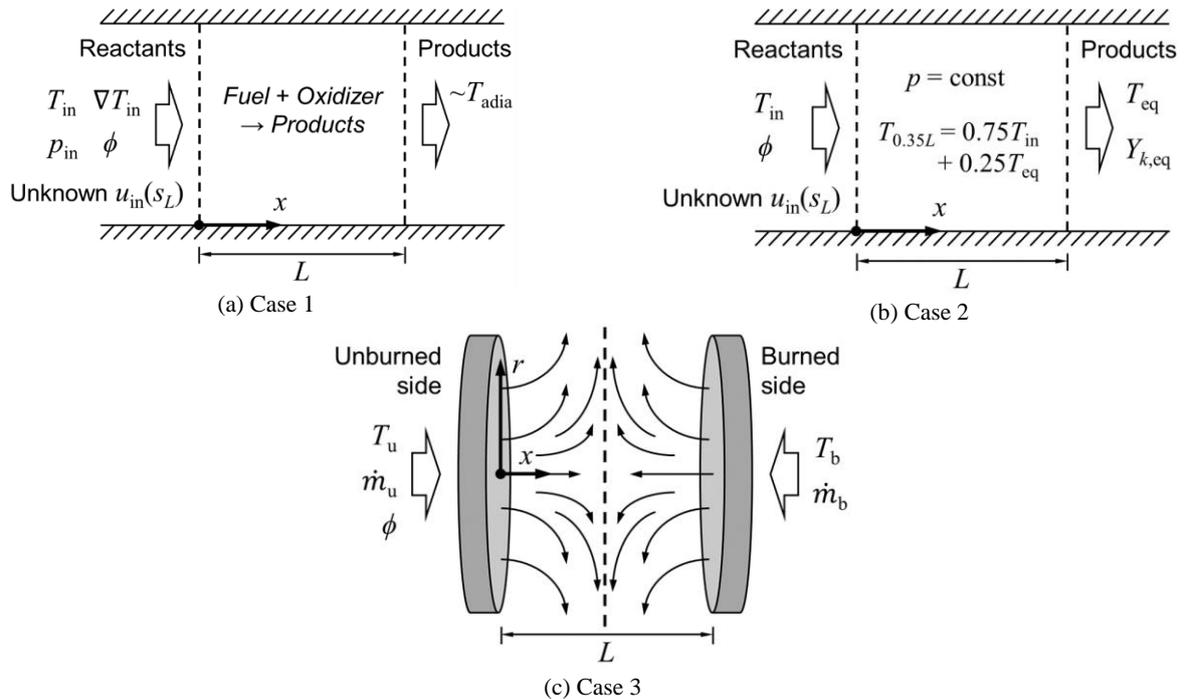

**Fig. 1.** Schematic of three test cases in this work.

## 2.1 Model and configurations of Case 1

Case 1 is a 1D FPP flame based on simplified models assuming constant material properties, unit Lewis number (*Le*), ideal gas, and inviscid flow. A 1-step irreversible reaction is assumed: *Fuel + Oxidizer → Product*. The fuel is assumed to be so lean that the reaction rate ($\omega$) can be calculated by:

$$\omega = A e^{-\frac{E_a}{RT}} (\rho Y_F)^\nu \tag{1}$$

where $A$ is the pre-exponential factor, $E_a$ is the activation energy, $R$ is the universal gas constant, $T$ is the temperature, $\rho$ is the gas density, $Y_F$ is the mass fraction of the fuel, and $\nu$ is the reaction order. The governing ODEs are:

$$\begin{cases} \dfrac{d(\rho u)}{dx} = 0 \\ \dfrac{d(\rho u u)}{dx} = -\dfrac{dp}{dx} \\ \rho u c_p \dfrac{dT}{dx} - \lambda \dfrac{d^2 T}{dx^2} = \omega q_F \\ \rho u \dfrac{d Y_F}{dx} - \rho D \dfrac{d^2 Y_F}{dx^2} = -\omega \end{cases} \tag{2}$$

where $u$ and $p$ represent the flow velocity and pressure, respectively. The $c_p$, $\lambda$, $q_F$, and $D$ are the heat capacity, thermal conductivity, fuel calorific value, and diffusion coefficient, respectively. The continuity equation can be reduced to $\rho u = \rho_{in} u_{in} = \rho_{in} s_L$, where $s_L$ is the laminar flame speed and is an eigenvalue to be solved together with the ODE solving. Based on the ideal gas equation ($pW = \rho RT$) and the unity *Le* assumption ($\lambda = \rho c_p D$), the 4-equation ODEs can be reduced to a 1-equation ODE:

$$\rho_{in} s_L c_p \frac{dT}{dx} - \lambda \frac{d^2 T}{dx^2} = \omega q_F \tag{3}$$

with the supplementary equations:

$$\begin{cases} u = \dfrac{c - \sqrt{c^2 - 4RT/W}}{2}, \text{ where } c = s_L + \dfrac{RT_{in}}{W s_L} \\ \rho = \dfrac{\rho_{in} s_L}{u} \\ Y_F = Y_{F,in} + \dfrac{c_p (T_{in} - T)}{q_F} \end{cases} \tag{4}$$

where $W$ is the gas molecular weight. Therefore, all variables ($\rho$, $u$, $p$, $Y_F$, $\omega$) are functions of $T$. The parameters involved in the ODEs are listed in Table 1.

**Table 1**

Parameters of Case 1, using SI units.

| Parameter | Symbol | Value |
| --- | --- | --- |
| Universal gas constant | $R$ | 8.3145 |
| Pre-exponential factor | $A$ | $1.4\times10^8$ |
| Reaction order | $\nu$ | 1.6 |
| Activation energy | $E_a$ | 121417.2 |
| Molecular weight | $W$ | 0.02897 |
| Thermal conductivity | $\lambda$ | 0.026 |
| Heat capacity | $c_p$ | 1000.0 |
| Fuel calorific value | $q_F$ | $5.0\times10^7$ |

Case 1 involves two BCs: the inlet value and the outlet vanishing gradient of $T$. The latter is not explicitly given since it is satisfied as long as the domain length ($L$) is sufficient. The inlet gradient of $T$ is also provided to offset the degree of freedom introduced by the unknown inlet velocity ($s_L$) and to avoid the "cold boundary" problem. We set $T_{in}$ = 298 K, $(dT/dx)_{in}$ = $10^5$ K/m, and $L$ = 1.5 mm. In addition, the inlet pressure $p_{in} \in [1.0, 1.2]$ atm, the inlet equivalence ratio $\phi \in [0.4, 0.42, 0.44, 0.46, 0.48, 0.5]$, considering only lean fuel conditions to satisfy the model assumptions. The fuel mass fraction at the inlet $Y_{F,\,in} = \phi / (4 + \phi)$, which assumes a methane-oxygen ($CH_4$-$O_2$) reaction.

The reference solution of Case 1 is obtained by FDM with a first-order Euler scheme and 10000 grids, and $s_L$ is calculated using the bisection method. The criteria for updating $s_L$ are as follows: $T > T_{adia}$, it indicates flame blowing out and an excessively large $s_L$, whereas if $dT/dx < 0$, it indicates flame flashback and an insufficient $s_L$. The convergence criterion is that the bisection interval must be smaller than $10^{-16}$ m/s. The adiabatic flame temperature ($T_{adia}$) is obtained by setting $Y_F = 0$. The detailed algorithm is provided in Appendix A.

## 2.2 Model and configurations of Case 2 and Case 3

Case 2 and Case 3 are 1D FPP and CFP flames, respectively, where more practical models are considered. Their unified governing ODEs are [4, 70]:

$$\begin{cases} \dfrac{d\rho u}{dx} + 2\rho V = 0 \\ \rho u \dfrac{dV}{dx} + \rho V^2 = -\Lambda + \dfrac{d}{dx}\left(\mu \dfrac{dV}{dx}\right) \\ \rho c_p u \dfrac{dT}{dx} = \dfrac{d}{dx}\left(\lambda \dfrac{dT}{dx}\right) - \sum_{k=1}^{K} j_k \dfrac{dh_k}{dx} - \sum_{k=1}^{K} h_k W_k \dot{\omega}_k \\ \rho u \dfrac{dY_k}{dx} = -\dfrac{dj_k}{dx} + W_k \dot{\omega}_k, \quad k = 1, 2, \ldots, K \end{cases} \quad (5)$$

where $V = v/r$ is the spread rate, $\Lambda = (\partial p/\partial r)/r$ is the pressure curvature, and $\mu$ is the gas viscosity. The $h_k$, $j_k$, $W_k$, $\dot{\omega}_k$, and $Y_k$ are the enthalpy, diffusive mass flux, molecular weight, molar production rate, and mass fraction of the $k^{th}$ species, respectively. For FPP flames, $V = 0$ and $\Lambda = 0$, so the momentum equation vanishes and the continuity equation reduces to $\rho u = \rho_{in} u_{in} = \rho_{in} s_L$. $\Lambda$ and $s_L$ are eigenvalues for counterflow and FPP flames, respectively. A constant pressure is assumed and the mixture density is calculated by the ideal gas law:

$$\rho = \dfrac{pW}{RT} \quad (6)$$

where $W$ is the mean gas molecular weight:

$$W = \left(\sum_{k=1}^{K} Y_k / W_k\right)^{-1} \quad (7)$$

The thermodynamic properties, including the heat capacity, enthalpy, and entropy, are calculated using the NASA 7-coefficient polynomials (standard pressure, mole-based):

$$\begin{cases} C_{pk}^\circ / R = \sum_{n=0}^{4} a_{nk} T^n \\ H_k^\circ / (RT) = \sum_{n=0}^{4} a_{nk} T^n / (n+1) + a_{5k}/T \\ S_k^\circ / R = \sum_{n=1}^{4} a_{nk} T^n / n + a_{0k} \ln T + a_{6k} \end{cases} \quad (8)$$

At pressure $p$, the mass-based properties are:

$$\begin{cases} c_{pk} = C_{pk}/W_k = C_{pk}^\circ/W_k \\ h_k = H_k/W_k = H_k^\circ/W_k \\ s_k = S_k/W_k = \left[S_k^\circ - R\ln(p/p^\circ)\right]/W_k \end{cases} \qquad (9)$$

and the mean thermodynamic properties are:

$$\begin{cases} c_p = \sum_{k=1}^{K} c_{pk} Y_k \\ h = \sum_{k=1}^{K} h_k Y_k \\ s = S/W, \text{ where } S = \sum_{k=1}^{K}\left(S_k - R\ln X_k\right)X_k \end{cases} \qquad (10)$$

where $X_k$ are the molar fractions of the $k^{\text{th}}$ species:

$$X_k = \frac{Y_k W}{W_k} \qquad (11)$$

The transport properties, including the viscosity ($\mu_k$), thermal conductivity ($\lambda_k$), and binary diffusion coefficients ($D_{kj}$), are also temperature dependent:

$$\begin{cases} \mu_k^{1/2} = T^{1/4} \sum_{n=0}^{4} b_{n,k} (\ln T)^n \\ \lambda_k = T^{1/2} \sum_{n=0}^{4} c_{n,k} (\ln T)^n \\ D_{kj} = T^{3/2} \sum_{n=0}^{4} d_{n,kj} (\ln T)^n \end{cases} \qquad (12)$$

Based on the mixture-averaged model, the mean transport properties are:

$$\mu = \sum_{k=1}^{K} \frac{\mu_k X_k}{\sum_{j=1}^{K} \Phi_{kj} X_j}, \text{ where } \Phi_{kj} = \frac{1}{\sqrt{8}}\left(1+\frac{W_k}{W_j}\right)^{-\frac{1}{2}}\left[1+\left(\frac{\mu_k}{\mu_j}\right)^{\frac{1}{2}}\left(\frac{W_j}{W_k}\right)^{\frac{1}{4}}\right]^2 \qquad (13)$$

$$\lambda = \frac{1}{2}\left[\sum_{k=1}^{K} X_k \lambda_k + \left(\sum_{k=1}^{K} X_k/\lambda_k\right)^{-1}\right] \qquad (14)$$

$$D'_{km} = \frac{1-Y_k}{\sum_{j \neq k}^{K} X_j/D_{jk}} \qquad (15)$$

The species diffusive mass fluxes ($j_k$) are then computed by:

$$\begin{aligned} V'_k &= -\frac{1}{X_k} D'_{km} \frac{dX_k}{dx} \\ j_k^* &= \rho Y_k V'_k \\ j_k &= j_k^* - Y_k \sum_{i=1}^{K} j_i^* \end{aligned} \qquad (16)$$

where $V_k'$ are the diffusion velocities. The last equation is a correction in the mixture-averaged model to make $j_k$ satisfy the condition of zero net species diffusive flux.

For the chemical reaction, we adopted the 1S_CH4_MP1 mechanism [70], which is a 1-step irreversible mechanism of $CH_4$-air reaction: $CH_4 + 2O_2 \rightarrow CO_2 + 2H_2O$, where 5 species (including $N_2$) are considered and the reaction order of $O_2$ is 0.5. Therefore, the reaction source terms ($\dot{\omega}_k$) are computed by:

$$k_f = AT^b \exp\left(-\frac{E_a}{RT}\right) \qquad (17)$$

$$q = k_f [X_{CH4}][X_{O2}]^{0.5} \qquad (18)$$

$$\dot{\omega}_k = v'_k q \tag{19}$$

where $k_f$ is the forward rate constant, $q$ is the rate-of-progress variable, $[X_k]$ are the molar concentrations:

$$[X_k] = \frac{\rho Y_k}{W_k} \tag{20}$$

The pre-exponential factor, temperature exponent, and activation energy are: $A$ = 1.1e7 m$^{1.5}$/(mol$^{0.5}$·s), $b$ = 0, $E_a$ = 83680 J/mol. The forward stoichiometric coefficients $v_k'$ are −1, −2, 1, 2, 0 for $CH_4$, $O_2$, $CO_2$, $H_2O$, and $N_2$, respectively.

The reference solutions of Case 2 and Case 3 are obtained by Cantera [70], an open-source suite of tools for various combustion problems, which adopted the aforementioned models. Default solution settings and 501 uniform girds are adopted. For Case 2, $T_{in}$ = 300K, $p \in$ [1, 5] atm, $\phi \in$ [0.6, 0.8, 1.0, 1.2, 1.4, 1.6]. The domain length $L$ = 2.5 and 1 mm for $p$ = 1 and 5 atm, respectively. Different from the practice of Case 1, to offset the degree of freedom brought by $s_L$, a fixed temperature condition is added: $T_{0.35L} = 0.75 T_{in} + 0.25 T_{eq}$, which is the default setting of Cantera. The equilibrium temperature $T_{eq}$ (= $T_{outlet}$) can be easily calculated before solving the ODEs. For Case 3, $p$ = 1 atm, $T_{in,u} = T_{in,b}$ = 300K (u: unburned; b: burned). The mass flow rates $\dot{m}_u = \dot{m}_b$ = 1 kg/(m$^2$·s). The unburned side is $CH_4$-air mixture and the burned side is $N_2$. The nozzle distance $L$ = 2 cm.

## 2.3 FlamePINN-1D framework

Fig. 2 shows the schematic of the FlamePINN-1D framework for solving the ODE problems of 1D laminar flames (the yellow part), which consist of ODEs, BCs, and observation conditions (OCs), given by

$$\mathcal{F}_i \left[ \mathbf{u}(x); \boldsymbol{\kappa} \right] = 0, \ x \in \Omega, \ i = 1, 2, ..., N_{\text{ODE}} \tag{21}$$

$$\mathcal{B}_i \left[ \mathbf{u}(x); \boldsymbol{\kappa} \right] = 0, \ x \in \partial\Omega, \ i = 1, 2, ..., N_{\text{BC}} \tag{22}$$

$$\mathcal{O}_i \left[ \mathbf{u}(x); \boldsymbol{\kappa} \right] = 0, \ x \in \Omega, \ i = 1, 2, ..., N_{\text{OC}} \tag{23}$$

where $\mathcal{F}$, $\mathcal{B}$, and $\mathcal{O}$ are the corresponding operators, $\Omega$ and $\partial\Omega$ are the calculation domain and its boundaries, and $\mathbf{u}$ and $\boldsymbol{\kappa}$ are the vectors of dependent variables and system parameters (such as the eigenvalues and material properties). Note that OCs exist only in inverse problems, where the observations are sparse and may be partial, i.e., not all components of $\mathbf{u}$ are observed. Additionally, for inverse problems, BCs may be incomplete and $\boldsymbol{\kappa}$ may be unknown. Based on the observations, field reconstruction means predicting the continuous field, and parameter inference means inferring the unknown parameters ($\boldsymbol{\kappa}$).

To solve the ODE problem, FlamePINN-1D first uses an NN to approximate the solution, denoted by $\mathbf{u_\theta}^\# =$ NN($x^\#$; $\boldsymbol{\theta}$), where $x^\#$ is the transformed $x$-coordinate, $\mathbf{u_\theta}^\#$ is the predicted vector of transformed dependent variables, and $\boldsymbol{\theta}$ are the trainable parameters of the NN. Here, the transformations are usually normalization or non-dimensionalization, in which case we will denote $x^\#$ and $\mathbf{u_\theta}^\#$ as $x^*$ and $\mathbf{u_\theta}^*$ respectively. The mapping between $x$ and $\mathbf{u_\theta}$ can be represented by: $\mathbf{u_\theta}$ = Map($x$; $\boldsymbol{\theta}$). Based on AD, the derivatives of $\mathbf{u_\theta}$ with respect to $x$ can be obtained and substituted into $\mathcal{F}$, $\mathcal{B}$, and $\mathcal{O}$ to get the corresponding residuals. Other variables, such as density and transport properties, can also be calculated based on $T$ and $Y$ and substituted into $\mathcal{F}$, $\mathcal{B}$, and $\mathcal{O}$. The loss terms are the mean square errors (MSEs) of each residual:

$$\mathcal{L}_{\text{ODE}i}(\boldsymbol{\theta}, \boldsymbol{\kappa}; \mathcal{T}_{\text{ODE}i}) = \frac{1}{|\mathcal{T}_{\text{ODE}i}|} \sum_{x_j \in \mathcal{T}_{\text{ODE}i}} \left| \mathcal{F}_i \left[ \mathbf{u_\theta}(x_j); \boldsymbol{\kappa} \right] \right|^2, \ \mathcal{T}_{\text{ODE}i} := \{x_j \mid x_j \in \Omega\}_{j=1}^{|\mathcal{T}_{\text{ODE}i}|} \tag{24}$$

$$\mathcal{L}_{\text{BC}i}(\boldsymbol{\theta}, \boldsymbol{\kappa}; \mathcal{T}_{\text{BC}i}) = \frac{1}{|\mathcal{T}_{\text{BC}i}|} \sum_{x_j \in \mathcal{T}_{\text{BC}i}} \left| \mathcal{B}_i \left[ \mathbf{u_\theta}(x_j); \boldsymbol{\kappa} \right] \right|^2, \ \mathcal{T}_{\text{BC}i} := \{x_j \mid x_j \in \partial\Omega\}_{j=1}^{|\mathcal{T}_{\text{BC}i}|} \tag{25}$$

$$\mathcal{L}_{\text{OC}i}(\boldsymbol{\theta},\boldsymbol{\kappa};\mathcal{T}_{\text{OC}i}) = \frac{1}{|\mathcal{T}_{\text{OC}i}|}\sum_{x_j \in \mathcal{T}_{\text{OC}i}} \left|\mathcal{O}_i\left[\mathbf{u}_{\boldsymbol{\theta}}(x_j);\boldsymbol{\kappa}\right]\right|^2, \quad \mathcal{T}_{\text{OC}i} := \left\{(x_j, u_{i,j}) \mid x_j \in \Omega\right\}_{j=1}^{|\mathcal{T}_{\text{OC}i}|} \tag{26}$$

where $\mathcal{T}_{\text{ODE}i}$ and $\mathcal{T}_{\text{BC}i}$ are the coordinate point sets sampled on the domain and its boundaries, respectively, and $\mathcal{T}_{\text{OC}i}$ is the observed point set for component $u_i$. This shows that PINN is indeed a mesh-free method. The final loss function is the weighted sum of all loss terms:

$$\mathcal{L}(\boldsymbol{\theta},\boldsymbol{\kappa}) = \sum_{i=1}^{N_{\text{ODE}}+N_{\text{BC}}+N_{\text{OC}}} w_i \mathcal{L}_i(\boldsymbol{\theta},\boldsymbol{\kappa}) \tag{27}$$

The NN parameters and unknown system parameters are updated using gradient-based optimization algorithms:

$$\boldsymbol{\theta}_{k+1} = \boldsymbol{\theta}_k - \eta_k f(\nabla_{\boldsymbol{\theta}}\mathcal{L}_k) \tag{28}$$

$$\boldsymbol{\kappa}_{k+1} = \boldsymbol{\kappa}_k - \eta_k f(\nabla_{\boldsymbol{\kappa}}\mathcal{L}_k) \tag{29}$$

where $\eta_k$ is the learning rate at the $k^{\text{th}}$ iteration and $f$ is a function specified by the optimizer. The training will stop after predefined iterations/epochs.

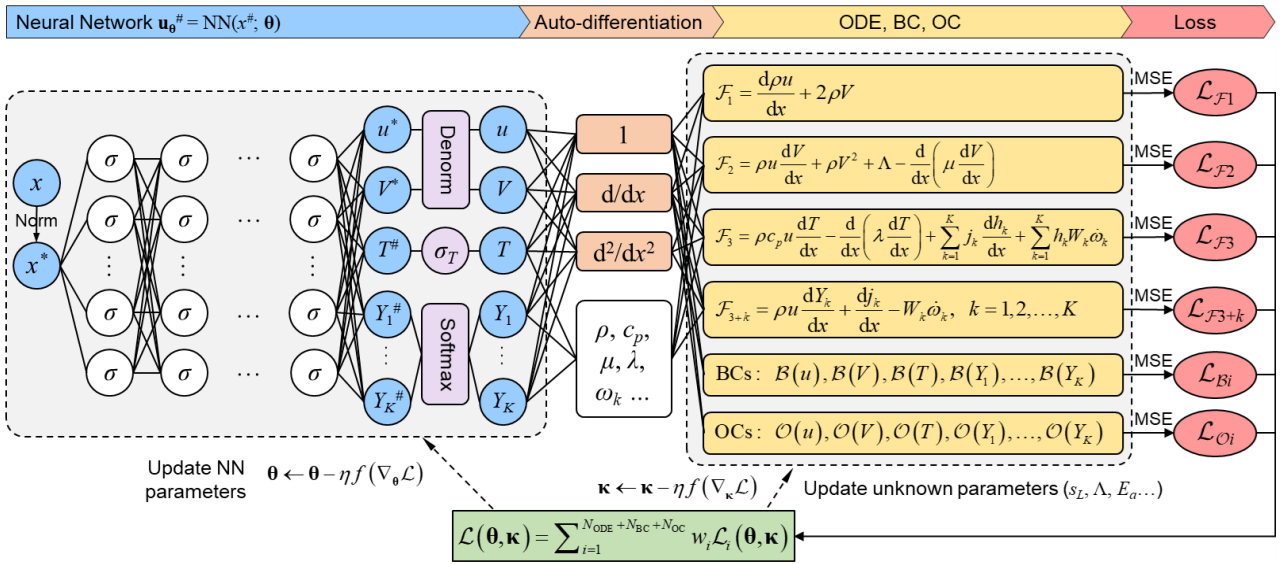

**Fig. 2.** Schematic of FlamePINN-1D framework for solving forward and inverse problems of 1D laminar flames based on the complex model. BC: boundary condition. OC: observation condition, which is absent in forward problems. The star superscript represents normalized variables. For FPP flames, the NN doesn't output $u$ and $V$, and $\mathcal{F}_1$ and $\mathcal{F}_2$ are absent. For the simplified model (Case 1), only $T$ is output by the NN and the ODE part is different. In practice, the ODEs, BCs, and OCs of normalized variables instead of the original ones are adopted.

### 2.4 Transformation of variables in FlamePINN-1D

The appropriate transformation of the input and output variables often plays an essential role in the robust training of NNs. For the cases in this work, the transformations of $x$, $u$, and $V$ are normalizations: $x^* = \alpha_x(x + \Delta_x)$, $u^* = \alpha_u u$, $V^* = \alpha_V V$, which are the same as in our previous work [46]. For $T$, its range is constrained to avoid unphysical results:

$$\begin{aligned} T &= \text{Sigmoid}(T^{\#}) \cdot (T_{\max} - T_{\min}) + T_{\min} \\ T^* &= \alpha_T T \end{aligned} \tag{30}$$

where $\text{Sigmoid}(x) = (1+e^{-x})^{-1}$. The species mass fractions are transformed by:

$$Y_k = \text{Softmax}\left(Y_k^{\#}\right) = \frac{\exp\left(Y_k^{\#}\right)}{\sum_{k=1}^{K}\exp\left(Y_k^{\#}\right)} \tag{31}$$

$$Y_k^* = \alpha_{Yk} Y_k$$

so that the range constraint ($0 \leq Y_k \leq 1$) and the species conservation ($\sum_k Y_k = 1$) can be exactly satisfied. Species conservation can also be constrained by adding a loss term [63], which is a soft rather than a hard constraint.

Note that in practice, unlike in Fig. 2, the ODEs, BCs, and OCs of normalized variables are adopted instead of the original ones [46], following the preference of NNs to avoid gradient vanishing or explosion. For example, $\mathcal{B}^*(T) = \alpha_T \mathcal{B}(T)$ is used instead of $\mathcal{B}(T)$. For the ODEs, the ODEs of normalized variables can be derived [46, 57], which does not need to be implemented explicitly, since the transformed ODEs are usually just multiples of the original ones. Therefore, we can still take advantage of the convenience of AD to calculate $\mathcal{F}$ directly and convert it to $\mathcal{F}^*$ manually. As in Ref. [46], the scales of $\mathcal{F}$ are based on the main term of $\mathcal{F}$. Specifically, $\mathcal{F}_1^* = (\alpha_u/\alpha_x)\mathcal{F}_1$, $\mathcal{F}_2^* = (\alpha_u\alpha_V/\alpha_x)\mathcal{F}_2$, $\mathcal{F}_3^* = (\alpha_u\alpha_T/1000\alpha_x)\mathcal{F}_3$, and $\mathcal{F}_{3+k}^* = (\alpha_u\alpha_{Yk}/\alpha_x)\mathcal{F}_{3+k}$, where the "1000" is to scale $c_p$. Additionally, the unknown physics-related parameters ($\boldsymbol{\kappa}$) should also be scaled to an appropriate order of magnitude (OoM) [46], which is $\mathcal{O}(0.1)$–$\mathcal{O}(1)$ in this work, by $\boldsymbol{\kappa}^* = \alpha_{\boldsymbol{\kappa}}\boldsymbol{\kappa}$.

The parameters of the transformations are listed in Table 2. The maximum values of the dependent variables are scaled to 5 so that their OoMs are $\mathcal{O}(1)$ [46]. Our previous work has validated that local normalization, rather than global normalization, should be adopted for thin-layer problems [46], so the domain lengths are scaled to greater than 1 to mitigate the thin-layer challenges. For Case 2 and Case 3, the domains are also shifted so that the thin layers can match the linear interval of the activation functions [46]. For Case 1, as mentioned above, $T > T_{\text{adia}}$ implies flame blowing out, so the range constraint of $T$ is also a hard constraint of not blowing out. For Case 3, since only inverse problems are solved (details in the next section), the scales are calculated based on the observed data.

**Table 2**

Parameters of the variable transformations. The $\alpha_\lambda$ and $\alpha_{Ea}$ of Case 1 are only valid for the inverse problem. The subscript "ob" means observation.

| Parameter | Case 1 | Case 2 | Case 3 |
|---|---|---|---|
| $\alpha_x$ | $10/L$ | $4/L$ | $4/L$ |
| $\Delta_x$ | 0 | $-L/2$ | $-L/2$ |
| $\alpha_u$ | – | – | $5/\max(\|u_{\text{ob}}\|)$ |
| $\alpha_V$ | – | – | $5/\max(\|V_{\text{ob}}\|)$ |
| $\alpha_T$ | $5/T_{\text{adia}}$ | $5/T_{\text{eq}}$ | $5/\max(T_{\text{ob}})$ |
| $\alpha_{Yk}$ | – | $5/\max(Y_{k,\text{in}}, Y_{k,\text{eq}})$ | $5/\max(Y_{k,\text{ob}})$ |
| $T_{\min}$ | $T_{\text{in}}$ | $T_{\text{in}}$ | $\min(T_u, T_b)$ |
| $T_{\max}$ | $T_{\text{adia}}$ | $T_{\text{eq}}$ | 3000K |
| $\alpha_{\boldsymbol{\kappa}}$ | $\alpha_{sL} = 1$, $\alpha_\lambda = 10$, $\alpha_{Ea} = 10^{-6}$ | $\alpha_{sL} = 1$ | $\alpha_\Lambda = 10^{-4}$, $\alpha_{Ea} = 10^{-5}$ |

*2.5 Problem and PINN configurations*

The PINN configurations in this work are listed in Table 3. For Case 1, both forward and inverse problems are studied. For the forward problem, the BCs include the inlet value of $T$ and its gradient (see Section 2.1). For the inverse problem, the BCs are absent and we assume that the thermal conductivity ($\lambda$) and activation energy ($E_a$) are unknown. We use FlamePINN-1D to learn these parameters along with $s_L$ from uniformly distributed noisy sparse observations of $T$ and $u$. For Case 2, only forward problems are studied, where the BCs include the inlet and outlet BCs of $T$ and $Y$, as well as the fixed temperature condition ($T_{0.35L} = 0.75T_{\text{in}} + 0.25T_{\text{eq}}$). For Case 3, only the inverse problems are studied, including field reconstruction and parameter inference. Unlike Case 1, only $E_a$ is set to be learned since the material properties are not constants. The default initial values of the unknown parameters ($\kappa_{ini}$) are shown in Table 3, which are set based on the approximate OoMs of the corresponding parameters, which is a low-demanding prior knowledge.

**Table 3**

Default configurations of FlamePINN-1D in this work. The $\lambda_{ini}$ and $E_{a,ini}$ of Case 1 are only valid for the inverse problem. The first and last value of the NN size means the input and output dimension respectively, while "$A \times B$" means $B$ hidden layers and $A$ neurons per layer. The $\varphi$ represents $T$ for Case 1, while represents $T$ and $Y_k$ for Case 2.

| Content | Case 1 | Case 2 | Case 3 |
|---|---|---|---|
| Problem type | Both | Forward | Inverse |
| $\kappa_{ini}$ | $s_{L,ini} = 0.4$ m/s, $\lambda_{ini} = 0.03$ W/(m·K), $E_{a,ini} = 1.5\times10^5$ J/mol | $s_{L,ini} = 0.6$ m/s | $\Lambda_{ini} = -10^4$ Pa/m$^2$, $E_{a,ini} = 10^5$ J/mol |
| NN | MLP | MMLP-RWF | MMLP-RWF |
| NN size | [1, 64×3, 1] | [1, 64×6, 1+$K$] | [1, 64×6, 3+$K$] |
| $|\mathcal{T}_{ODE}|$ | 1001 | 100 | 100 |
| $\mathcal{T}_{ODE}$ sampling | Uniform. Fixed. | Random. Resample per 10 epochs. | Random. Resample per 10 epochs. |
| Warmup data | Z-curve by (0, $\varphi_{in}$), (0.3$L$, $\varphi_{in}$), (0.5$L$, $\varphi_{eq}$), ($L$, $\varphi_{eq}$) | Z-curve by (0, $\varphi_{in}$), (0.3$L$, $\varphi_{in}$), (0.5$L$, $\varphi_{eq}$), ($L$, $\varphi_{eq}$) | Observed data |
| Warmup stage 1 | 1000 epochs | 5000 epochs | 5000 epochs |
| Warmup stage 2 | None | 1000 epochs | None |
| Solving stage | 30000 epochs | 30000 epochs | 50000 epochs |
| $w_{ODE}$ | 1 | 1 | 1 |
| $w_{BC}$ or $w_{OC}$ | 1 | 1 | 100 |

For the NN architecture, a simple multi-layer perception (MLP) is used for Case 1, and a modified MLP (MMLP) [48, 71] with random weight factorization (RWF) [71, 72] is adopted for Case 2 and Case 3. The MLP formulation is:

$$\begin{aligned}
\mathbf{a}_0 &= x^{\#} \\
\mathbf{a}_l &= \sigma(\mathbf{z}_l), \ l = 1, 2, ..., L-1 \\
\mathbf{z}_l &= \mathbf{W}_l \mathbf{a}_{l-1} + \mathbf{b}_l, \ l = 1, 2, ..., L \\
\mathbf{u}_{\boldsymbol{\theta}}^{\#} &= \mathbf{z}_L
\end{aligned} \quad (32)$$

where $\mathbf{W}_l$ and $\mathbf{b}_l$ are weight matrices and biases of the $l^{th}$ layer. They are the trainable parameters ($\boldsymbol{\theta}$) of MLP. The activation function ($\sigma$) is set as *tanh* in all cases. The MMLP can be represented by:

$$\begin{aligned}
\mathbf{U} &= \sigma(\mathbf{W}_{\mathbf{U}} x^{\#} + \mathbf{b}_{\mathbf{U}}) \\
\mathbf{V} &= \sigma(\mathbf{W}_{\mathbf{V}} x^{\#} + \mathbf{b}_{\mathbf{V}}) \\
\mathbf{a}_0 &= x^{\#} \\
\mathbf{a}_l &= \sigma(\mathbf{z}_l) \odot \mathbf{U} + (1 - \sigma(\mathbf{z}_l)) \odot \mathbf{V}, \ l = 1, 2, ..., L-1 \\
\mathbf{z}_l &= \mathbf{W}_l \mathbf{a}_{l-1} + \mathbf{b}_l, \ l = 1, 2, ..., L \\
\mathbf{u}_{\boldsymbol{\theta}}^{\#} &= \mathbf{z}_L
\end{aligned} \quad (33)$$

where each hidden layer is reweighted by $\mathbf{U}$ and $\mathbf{V}$, which is similar to the attention mechanism. Compared to MLP, MMLP has more trainable parameters ($\mathbf{W}_{\mathbf{U}}$, $\mathbf{W}_{\mathbf{V}}$, $\mathbf{b}_{\mathbf{U}}$, and $\mathbf{b}_{\mathbf{V}}$) and a more complicated structure, leading to its better performance than MLP in many problems [71]. A schematic of MMLP can be found in Ref. [73]. RWF factorizes each weight matrix of an NN into the product of a vector ($\mathbf{s}$) and a new matrix ($\mathbf{M}$):

$$\mathbf{W} = \text{diag}(\mathbf{s}) \cdot \mathbf{M} \quad (34)$$

so that the trainable parameters become $\mathbf{s}$ and $\mathbf{M}$, which can shorten the optimization distance and yield better performance than PINNs without RWF.

As a mesh-free method, PINN samples coordinate data in the calculation domain to compute residuals, which can be done either uniformly or randomly. The numbers of ODE residual points ($|\mathcal{T}_{ODE}|$) are given in Table 3. For Case 2 and Case 3, a small amount of coordinate data is randomly resampled every 10 epochs to enhance model robustness [74] and alleviate computational burden. The observed points of the inverse problems will be discussed in the results section.

For FPP flames (Case 1 and Case 2), a warmup pretraining strategy is proposed to avoid unburned solutions. For Case 1, the NN is first trained to fit a Z-shaped curve of $T$ for a few epochs (warmup), and then trained for pure ODE solving, during which the Z-curve fitting loss is absent. This constitutes a two-stage training. For Case 2, an additional transitional training stage is added between these two stages, where both the Z-curve fitting and ODE solving losses are present. The Z-curve specification is given in Table 3, which is also the default initial solution in Cantera. Similarly, for the inverse problems of Case 3, a warmup pretraining is also implemented, where the observations serve as the warmup data. The validity of the warmup pretraining strategy and its mechanism will be discussed in the next section. The number of training epochs for each stage is given in Table 3. For all stages, the learning rate of the $k^{\text{th}}$ epoch is $\eta_k = 10^{-3} \times 0.95^{[k/1000]}$, where a decaying strategy is adopted and $[\cdot]$ denotes the floor function. The optimizers used are all Adam [75]. The loss weights ($w$) in Eq. (27) are hyperparameters that are often set empirically. The values of $w$ in Table 3 are found to yield sufficient performance. Dynamic loss weights [48, 49] can also be adopted in future work.

The computations are implemented based on the PINN library DeepXDE [76] and use Pytorch as the backend. To evaluate the model performance, the $L_2$ relative error ($L_2$RE) is used as the metric:

$$L_2\text{RE}_\varphi = \frac{\|\varphi_{\text{PINN}} - \varphi_{\text{ref}}\|_2}{\|\varphi_{\text{ref}}\|_2} = \frac{\sqrt{\sum_{i=1}^{N}(\varphi_{i,\text{PINN}} - \varphi_{i,\text{ref}})^2}}{\sqrt{\sum_{i=1}^{N}\varphi_{i,\text{ref}}^2}} \tag{35}$$

where $\varphi$ is a field variable and "ref" denotes the reference value. The number of uniformly distributed evaluation points ($N$) is 5001 for all cases. The relative error (RE) and its absolute value (absolute relative error, ARE) are used as metrics to evaluate the accuracy of inferred parameters.

## 3. Results

In this section, we first present and discuss the results of the forward problems (Case 1 and Case 2), followed by the results of the inverse problems (Case 1 and Case 3). Both quantitative and qualitative results are provided. Subsequently, the effectiveness of some adopted hyperparameters is validated through comparative studies. Finally, we discuss the computational efficiency of the FlamePINN-1D framework, focusing on the comparison between PINNs and traditional methods.

### 3.1 Forward problems
#### 3.1.1 Case 1 (simplified FPP flames)

Fig. 3 shows the predicted flame fields of PINN under two sets of conditions, where $T$ is predicted directly by PINN and other variables are calculated from $T$. The curves show excellent agreement between PINN and FDM for both weak and intense chemical reactions. At $p = 1$ atm and $\phi = 0.4$, the $L_2$RE of $T$ and $\omega$ are 0.13% and 1.08% respectively, while at $p = 1.2$ atm and $\phi = 0.5$, they are 0.32% and 3.21% respectively. Although the plots are not shown, at $p \in [1.0, 1.2]$ atm, $\phi \in [0.4, 0.42, 0.44, 0.46, 0.48, 0.5]$, the $L_2$RE of $T$, $Y_F$, $u$, $\rho$, and $p$ are all less than 2% and the $L_2$RE($\omega$) are all less than 5%, showing the great performance of PINN in solving the flame fields under various conditions. Although PINN involves some random operations, such as random initialization, the above conclusions about the $L_2$RE remain valid across a large number of independent runs.

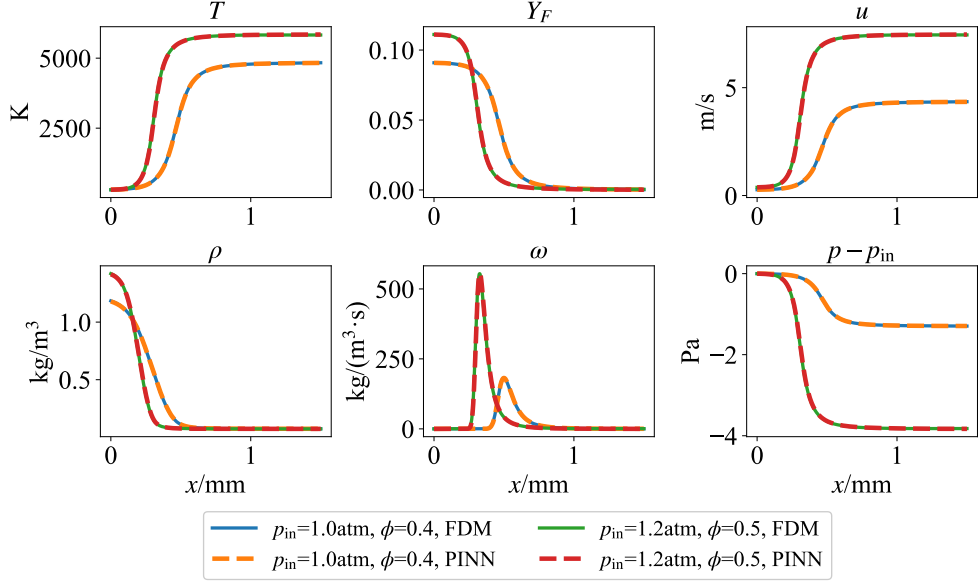

**Fig. 3.** Case 1 (forward problem): Comparison between the flame fields calculated by FDM and PINN under two sets of conditions, including the temperature ($T$), fuel mass fraction ($Y_F$), velocity ($u$), density ($\rho$), reaction rate ($\omega$), and relative pressure ($p - p_{in}$) fields. Only $T$ and $u_{in}$ ($s_L$) are directly predicted by PINN while other variables are derived based on them.

Fig. 4 and Fig. 5 give the results and learning curves of the laminar flame speeds, respectively. Almost no discrepancies can be observed between $s_L$ computed by the bisection iteration method and PINN, preliminarily demonstrating the feasibility of PINN to infer the eigenvalues in an ODE system of 1D laminar flames. Quantitatively, among these 12 $p$-$\phi$ conditions, the maximum absolute relative error between PINN and the bisection iteration method is only 0.03%. In Fig. 5, the inferred value of $s_L$ can converge to the reference value in the early training stage from different initial values, further illustrating the robustness of PINN in inferring $s_L$ under the initial tendency of both flashback and blowing out. The blowing-out initial values, which are larger than the true value, yield faster convergence than the flashback ones, suggesting the effectiveness of the designed hard constraint of not blowing out.

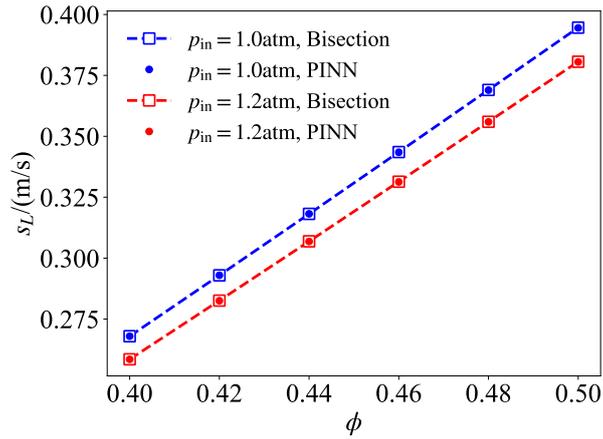

**Fig. 4.** Case 1 (forward problem): Comparison between the laminar flame speeds calculated by the bisection iteration and PINN. The initial value of $s_L$ is 0.4 m/s.

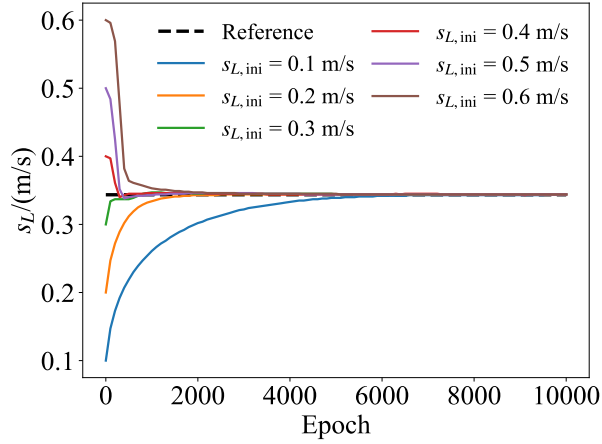

**Fig. 5.** Case 1 (forward problem): Learning histories of the laminar flame speed ($s_L$) from different initial values ($p_{in}$ = 1 atm, $\phi$ = 0.46). Only the early 10000 epochs are shown.

Fig. 6 compares the PINN performance without and with the warmup pretraining, showing that the predicted flame field can be completely unburned without warmup pretraining, where $T$ almost keeps the inlet value and the reaction rate is nearly zero throughout the entire domain (not shown). The loss history curves reveal that the mechanism of warmup pretraining to improve the performance of PINN is that it provides lower initial losses, so the losses can be reduced to quite low after the entire training process. From Fig. 6 (c), it can be observed that the more fundamental reason for the unburning solution is its high losses of BCs. Although the ODE residual is quite small, the residual of $(dT/dx)_{in}$ is quite high (compared to the good solution), which means that $(dT/dx)_{in}$ is too small and the energy flux is insufficient to burn the gas. In addition, the warmup pretraining took only a few seconds, which is less than 1% of the total training time, demonstrating its high efficiency.

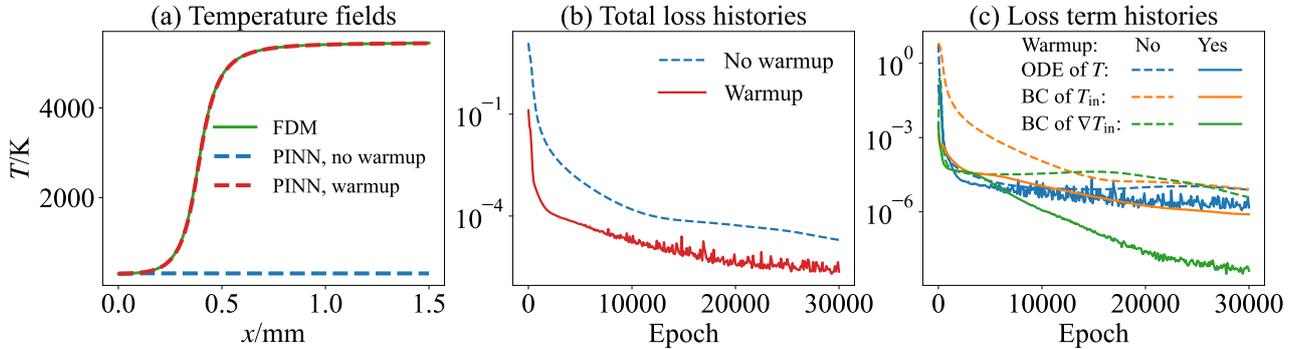

**Fig. 6.** Case 1 (forward problem): Comparison between PINN performance without and with warmup pretraining. (a) Temperature fields. (b) Histories of total loss. (c) Histories of each loss term.

### 3.1.2 Case 2 (FPP flames with detailed physical models)

Table 4 presents the evaluation metrics of the PINN results for the 12 working conditions of Case 2, listing the $L_2$RE of PINN-solved flame fields and the RE of PINN-inferred laminar flame speeds of a random run. Almost all of the errors are less than 1%, strongly demonstrating the accuracy of FlamePINN-1D in solving the flame fields and inferring the laminar flame speeds of FPP flames with detailed physical models. To visually demonstrate the performance of PINN, Fig. 7 compares the flame fields calculated by PINN and Cantera under two working conditions, where the $T$ and $Y_k$ fields are directly output from the NN and the other fields are calculated from them.

It can be observed that the PINN-predicted results agree well with Cantera, even under high pressure conditions, where the variables change drastically in the very thin flame layer, posing great challenges for PINN in reducing the ODE losses.

**Table 4**

Case 2 (forward problem): $L_2$RE of PINN-solved flame fields and RE of PINN-inferred laminar flame speeds ($s_L$). Without loss of generality, only the results of one random run are given.

| $p$ | $\phi$ | $T$ | $Y_{O2}$ | $Y_{CH4}$ | $Y_{CO2}$ | $\rho$ | $u$ | $s_L$ |
|---|---|---|---|---|---|---|---|---|
| | 0.6 | 0.26% | 0.18% | 0.33% | 0.33% | 0.42% | 0.31% | 0.23% |
| | 0.8 | 0.16% | 0.17% | 0.28% | 0.17% | 0.33% | 0.14% | −0.04% |
| 1 atm | 1.0 | 0.53% | 0.45% | 0.41% | 0.44% | 0.51% | 1.44% | 1.00% |
| | 1.2 | 0.20% | 0.25% | 0.27% | 0.20% | 0.42% | 0.30% | −0.40% |
| | 1.4 | 0.25% | 0.28% | 0.33% | 0.29% | 0.47% | 0.27% | −0.40% |
| | 1.6 | 0.25% | 0.38% | 0.22% | 0.27% | 0.44% | 0.44% | −0.59% |
| | 0.6 | 0.16% | 0.11% | 0.22% | 0.22% | 0.23% | 0.16% | −0.08% |
| | 0.8 | 0.23% | 0.21% | 0.28% | 0.21% | 0.57% | 0.17% | 0.18% |
| 5 atm | 1.0 | 0.35% | 0.38% | 0.35% | 0.27% | 0.80% | 0.34% | 0.42% |
| | 1.2 | 0.25% | 0.28% | 0.43% | 0.22% | 0.52% | 0.51% | −0.53% |
| | 1.4 | 0.29% | 0.32% | 0.32% | 0.23% | 0.57% | 0.31% | −0.37% |
| | 1.6 | 0.31% | 0.36% | 0.20% | 0.23% | 0.72% | 0.70% | −0.49% |

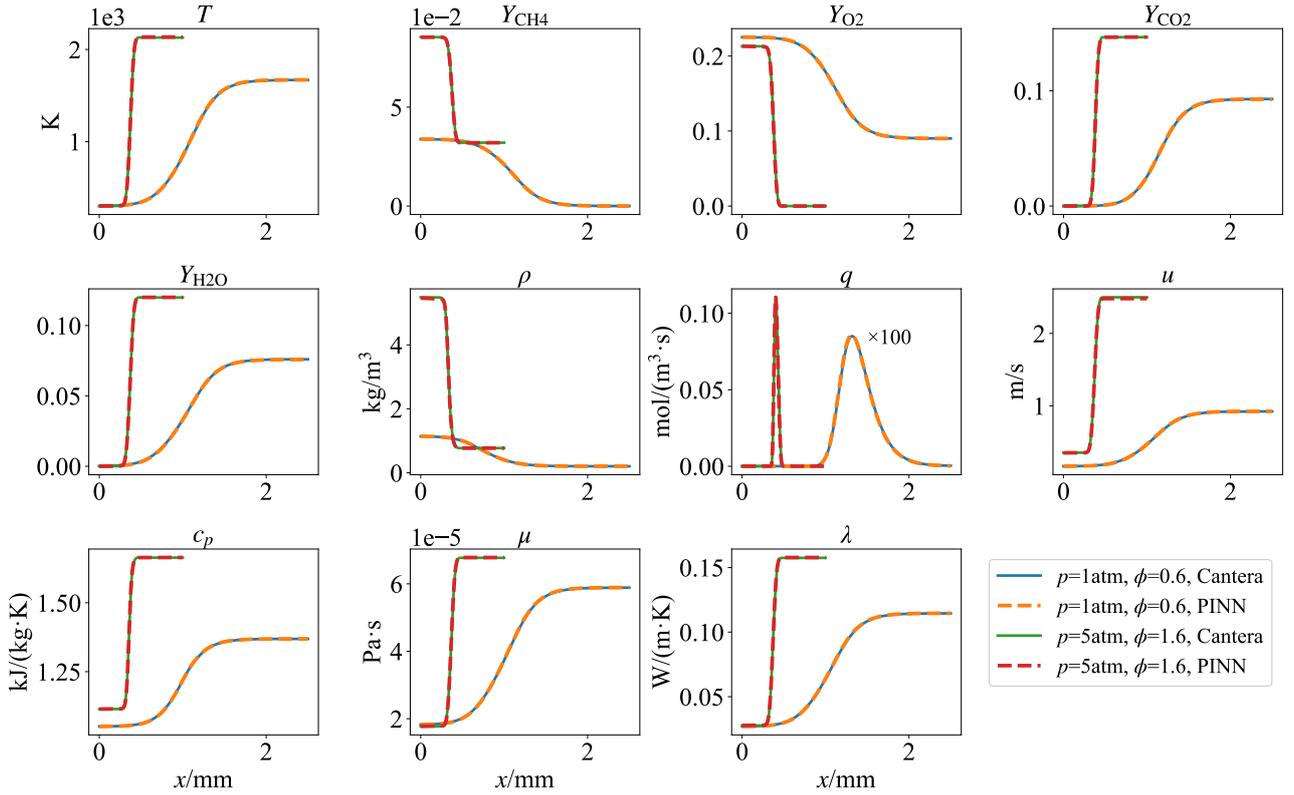

**Fig. 7.** Case 2 (forward problem): Comparison between the flame fields calculated by Cantera and PINN. The $T$ and $Y_k$ fields are directly output from the NN and the other fields are calculated from them.

For the laminar flame speeds, Fig. 8 shows the results of PINN-inferred laminar flame speeds under the 12 working conditions, whose REs are listed in Table 4. Again, the qualitative results show the accuracy of PINN in inferring the laminar flame speeds. Note that the laminar flame speeds should reach a maximum near $\phi = 1$, but do not in Fig. 8, which may be caused by the unphysical global reaction mechanism. This discrepancy is not our focus, as our primary aim is to prove the numerical effectiveness of PINN by comparing the results of Cantera and PINN based on the same physical models and working conditions. Fig. 9 plots the learning history curves of the laminar flame speed under some working conditions. With the same initial value, the optimization distances differ under different working conditions, which may pose challenges to PINNs. However, Fig. 9 shows that the inferred values can all eventually converge to the reference values, illustrating the robustness of FlamePINN-1D for parameter inference.

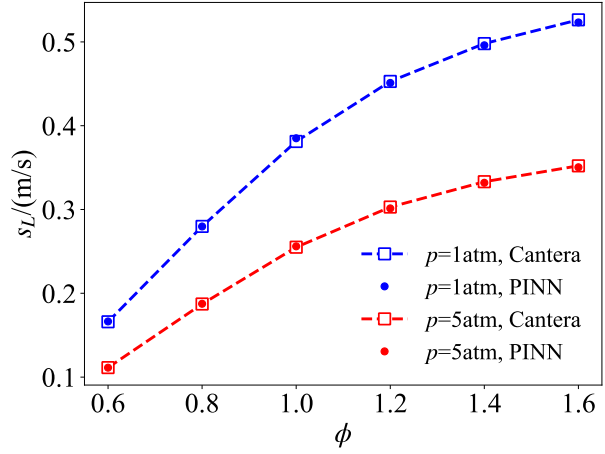

**Fig. 8.** Case 2 (forward problem): Comparison between the laminar flame speeds ($s_L$) calculated by Cantera and PINN.

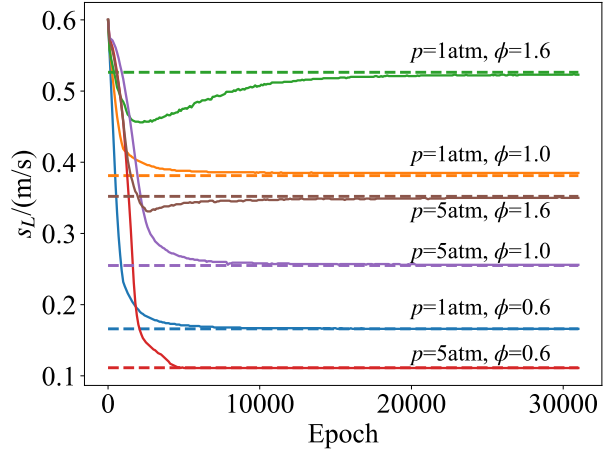

**Fig. 9.** Case 2 (forward problem): Learning histories of the laminar flame speed ($s_L$) under different working conditions. One color for one working condition. The dashed lines are the reference values obtained by Cantera under the corresponding working conditions. The learning consists of a transitional warmup stage of 1000 epochs and an ODE-solving stage of 30000 epochs.

The warmup pretraining plays the same role in the robust performance of PINN as in Case 1. Table 5 shows the results with different combinations of the warmup stages, and shows that PINN will fail to solve the problem without either of the two warmup stages, demonstrating the necessity of both warmup stages. Mathematically, the warmup stage 1 (Z-curve fitting) can quickly steer the result to a local optimum point, since the Z-curve is a good approximation of the true solution. However, the local optimum point indicated by the Z-curve fitting and the ODE residuals minimization may not be consistent, especially for Case 2, which is more difficult than Case 1. Therefore, if we omit the warmup stage 2 and directly transition from Z-curve fitting to ODE and BC residuals minimization, the result will jump out of the previous local optimum point, negating the efforts of stage 1. Thus, a transitional stage is necessary where the presence of the Z-curve fitting loss can mitigate this undesirable tendency.

**Table 5**

Case 2 (forward problem): $L_2$RE of PINN-solved flame fields and RE of PINN-inferred laminar flame speeds ($s_L$) with different combinations of the warmup stages. Stage 1 is the Z-curve fitting stage, while stage 2 is a transitional stage that minimizes both the Z-curve fitting loss and the ODE losses. The $p = 5$ atm and $\phi = 1.6$.

| Stage 1 | Stage 2 | $T$ | $Y_{CH4}$ | $u$ | $s_L$ |
|---|---|---|---|---|---|
| Yes | Yes | 0.31% | 0.20% | 0.70% | −0.49% |
| Yes | No | 25.17% | 26.86% | 36.81% | 26.21% |
| No | Yes | 34.54% | 19.05% | 95.29% | −93.25% |
| No | No | 33.59% | 20.44% | 95.37% | −93.44% |

## *3.2 Inverse problems: Field reconstruction and parameter inference*

### *3.2.1 Case 1 (simplified FPP flames)*

Table 6 lists the metrics of the reconstructed flame fields and inferred $s_L$, $\lambda$, and $E_a$ from 15 clean and noisy observed points. It can be seen that the reconstruction and inference errors are quite small under clean observations. Under noisy conditions, the errors are larger but still satisfactory. The influence of noise is small for the inference of $s_L$ and $E_a$, but larger for $\lambda$. The reason may be that the noise has a greater effect on the temperature gradient, which is crucial for the inference of $\lambda$. Although simplified models are adopted, the results illustrate the feasibility of PINN to simultaneously learn the unknown eigenvalue, transport property, and reaction parameter from noisy sparse observed field data. Note that the number of observed points ($N_{ob} = 15$) is only 0.15% of the grid number of the FDM solution ($N_{grid} = 10000$), demonstrating the significant few-shot learning ability of PINN. Fig. 10 plots the learning history curves of the three parameters based on 10 observed points, showing that the learning converges at the early 40000$^{th}$ epoch. The noise only affects the converged value, not the epoch at which convergence occurs.

**Table 6**

Case 1 (inverse problem): $L_2$RE of PINN-reconstructed flame fields and ARE of PINN-inferred laminar flame speeds ($s_L$), thermal conductivities ($\lambda$), and activation energies ($E_a$) under observations with different noise levels. The number of observed points $N_{ob} = 15$. The $p_{in} = 1$ atm and $\phi = 0.46$. Each result gives the mean value of 5 independent runs.

| Variable | No noise | 1% noise | 2% noise |
|---|---|---|---|
| $T$ | 0.067% | 0.216% | 0.416% |
| $Y_F$ | 0.123% | 0.398% | 0.766% |
| $u$ | 0.078% | 0.412% | 0.540% |
| $\omega$ | 0.593% | 1.690% | 3.266% |
| $s_L$ | 0.034% | 0.295% | 0.286% |
| $\lambda$ | 0.019% | 0.948% | 3.136% |
| $E_a$ | 0.018% | 0.266% | 0.434% |

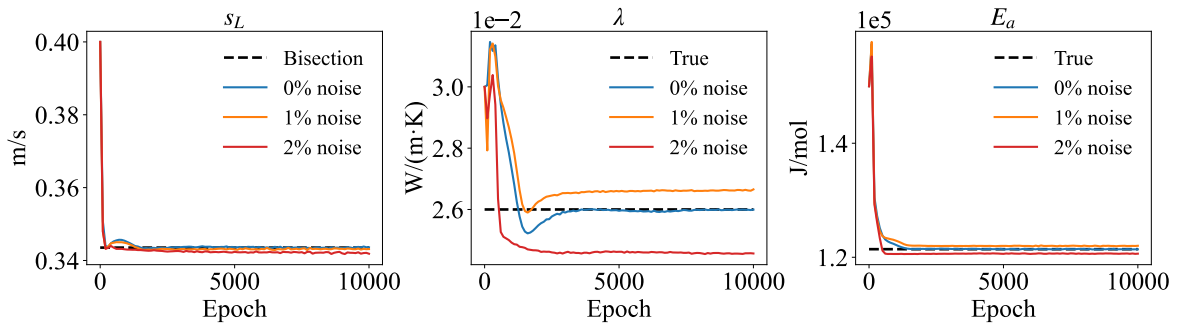

**Fig. 10.** Case 1 (inverse problem): Learning histories of laminar flame speed ($s_L$), thermal conductivity ($\lambda$), and activation energy ($E_a$) with different observation noise levels. The $p_{in} = 1$ atm, $\phi = 0.46$, and $N_{ob} = 10$. Only one random run is given.

To explore the sparsity limit of PINN for parameter inference, computations are performed with different numbers of observed points ($N_{ob}$), as shown in Fig. 11. The results show that under noise-free observations, the inference error can be quite small even $N_{ob}$ is as low as 5, indicating great few-shot learning ability of PINN. The noise poses challenges to the inference, especially for $N_{ob} < 10$ and for $\lambda$, but the errors of all parameters are quite small when $N_{ob} > 10$. Generally, the error decreases as $N_{ob}$ increases, but it is also affected by the observation distribution and the noise, so some nonlinear tendencies can be seen in Fig. 11. In short, the performance of PINN to infer the parameters from noisy sparse observations is validated, even only 0.1% of the reference values are observed. Additionally, the accurate and robust result of $E_a$ provides the possibility of learning chemical mechanisms from noisy sparse observations in more complex situations.

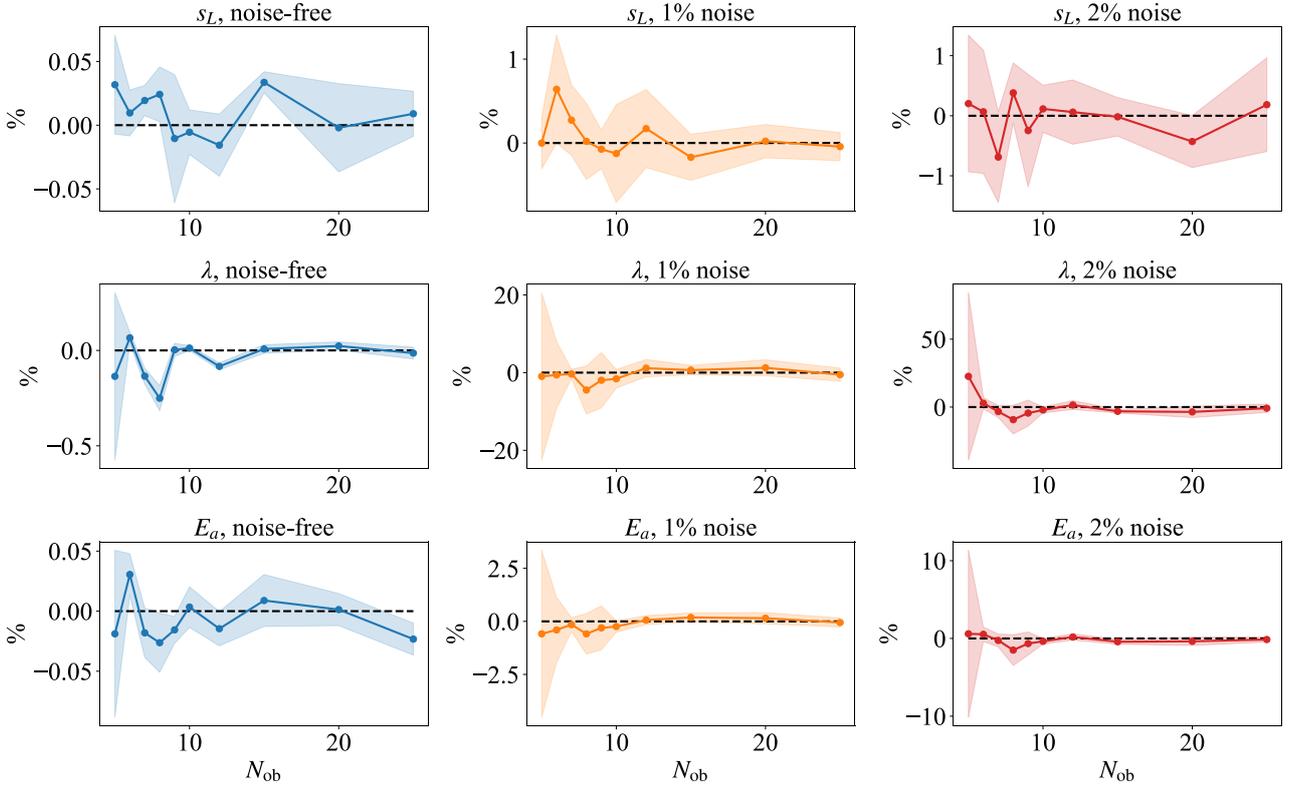

**Fig. 11.** Case 1 (inverse problem): RE of PINN-inferred laminar flame speed ($s_L$), thermal conductivity ($\lambda$), and activation energy ($E_a$) with different numbers of observed points ($N_{ob}$) and observation noise level. The $p_{in} = 1$ atm and $\phi = 0.46$. The solid lines are the mean values of 5 independent runs while the shaded areas represent the range of [Mean − Std, Mean + Std], where Std denotes the standard deviation.

*3.2.2 Case 3 (CFP flames with detailed physical models)*

Table 7 lists the metrics of PINN-reconstructed fields and PINN-inferred $\Lambda$ and $E_a$ from 16 observed points. For the field reconstruction, the $L_2$REs are all less than 5% for all variables and noise levels, quantitatively demonstrating the accuracy of PINN in reconstructing the flame fields with noisy sparse observations and detailed physical models. Since the reference solution uses 501 grid points, 16 points account for only 3.2% of them, indicating the observation sparsity. To show this more clearly, Fig. 12 compares the flame fields reconstructed by data fitting and PINN from fewer noisy observed points. The data fitting used the same NN and configurations as PINN, except that the ODE losses were absent. Obviously, PINN results match perfectly with the reference fields even with only 10 observed points, while data fitting fails to reconstruct the complex information smaller than the observational resolution, demonstrating that the physical constraint of the ODEs can effectively avoid overfitting. Moreover, unlike PINN, pure data fitting by NNs cannot infer the unknown system parameters.

**Table 7**

Case 3 (inverse problem): $L_2$RE of PINN-reconstructed flame fields and RE of PINN-inferred pressure curvatures ($\Lambda$) and activation energies ($E_a$). The number of observed points $N_{ob}$ = 16. Without loss of generality, only the results of one random run are given.

| Noise level | $\phi$ | $u$ | $V$ | $T$ | $Y_{O2}$ | $Y_{CH4}$ | $Y_{CO2}$ | $\rho$ | $\Lambda$ | $E_a$ |
|---|---|---|---|---|---|---|---|---|---|---|
| 0% | 0.7 | 2.60% | 0.41% | 2.54% | 1.58% | 2.30% | 3.68% | 2.23% | 13.84% | −0.08% |
| | 0.8 | 2.09% | 0.39% | 2.26% | 1.80% | 2.21% | 3.20% | 0.95% | 14.00% | −1.55% |
| | 0.9 | 4.03% | 0.53% | 2.90% | 3.07% | 3.41% | 3.74% | 3.26% | 14.28% | −0.44% |
| | 1.0 | 4.37% | 0.34% | 3.68% | 4.27% | 4.25% | 4.34% | 1.91% | 13.63% | −1.17% |
| | 1.1 | 1.36% | 0.14% | 1.05% | 1.63% | 1.44% | 1.53% | 1.79% | 14.04% | −1.88% |
| | 1.2 | 3.32% | 0.17% | 2.13% | 3.23% | 2.78% | 2.75% | 3.06% | 13.75% | −0.94% |
| 1% | 0.7 | 2.11% | 0.76% | 1.99% | 1.25% | 2.21% | 2.71% | 1.86% | 13.45% | 0.53% |
| | 0.8 | 2.03% | 1.11% | 2.36% | 1.77% | 2.25% | 3.47% | 1.00% | 13.99% | −1.30% |
| | 0.9 | 4.77% | 0.80% | 3.50% | 3.42% | 4.15% | 4.51% | 3.66% | 13.76% | −1.38% |
| | 1.0 | 4.87% | 0.97% | 3.97% | 4.70% | 4.54% | 4.98% | 2.08% | 13.27% | −1.36% |
| | 1.1 | 1.51% | 0.36% | 1.25% | 1.54% | 1.46% | 1.67% | 1.92% | 14.34% | −2.00% |
| | 1.2 | 2.74% | 1.40% | 1.96% | 3.55% | 3.00% | 2.47% | 2.57% | 14.33% | 0.83% |
| 2% | 0.7 | 4.29% | 1.56% | 4.64% | 2.53% | 3.50% | 5.79% | 3.22% | 13.96% | −2.81% |
| | 0.8 | 2.93% | 2.14% | 2.38% | 2.23% | 2.91% | 4.27% | 0.97% | 13.92% | −1.64% |
| | 0.9 | 4.61% | 0.90% | 3.64% | 3.57% | 4.43% | 4.94% | 3.81% | 14.32% | −1.46% |
| | 1.0 | 5.41% | 0.97% | 2.94% | 3.93% | 4.68% | 3.76% | 3.40% | 12.11% | 2.90% |
| | 1.1 | 3.12% | 1.40% | 1.64% | 2.33% | 1.77% | 2.62% | 2.07% | 15.36% | −2.72% |
| | 1.2 | 4.02% | 2.40% | 2.55% | 3.23% | 2.66% | 2.82% | 3.08% | 12.59% | −1.70% |

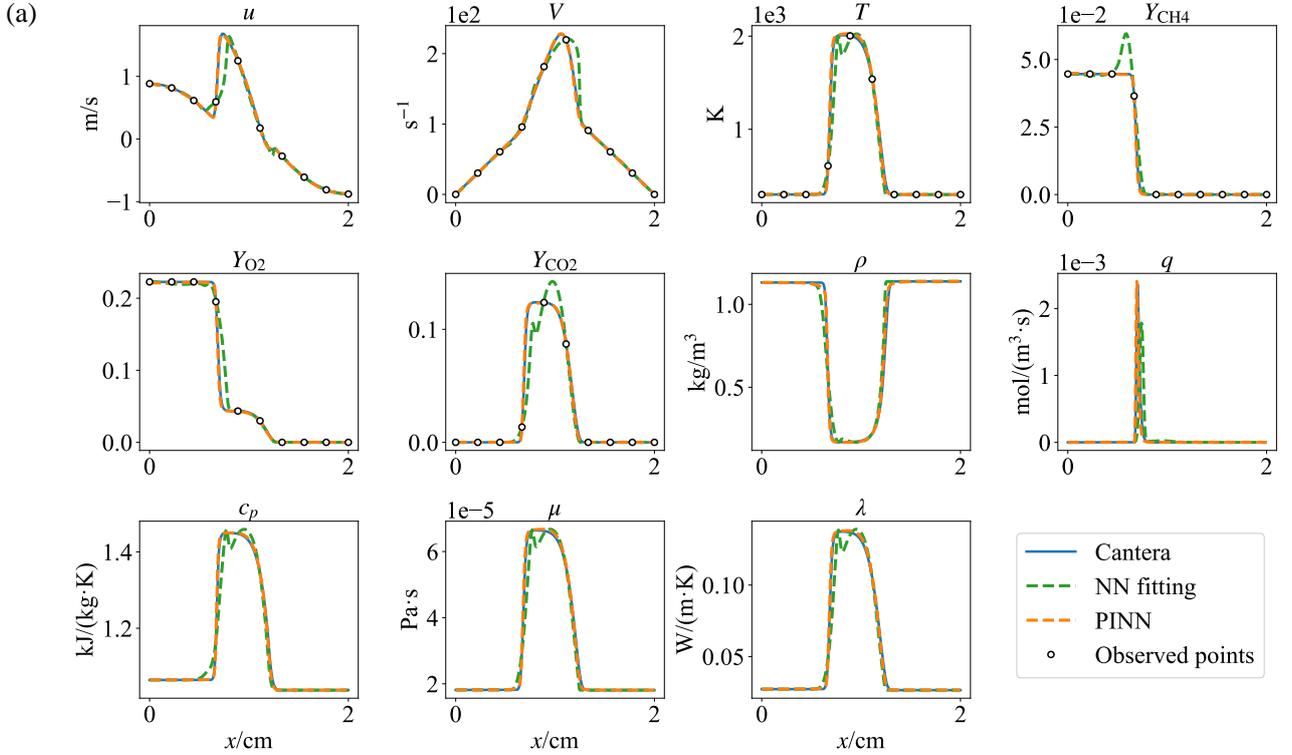

(a)

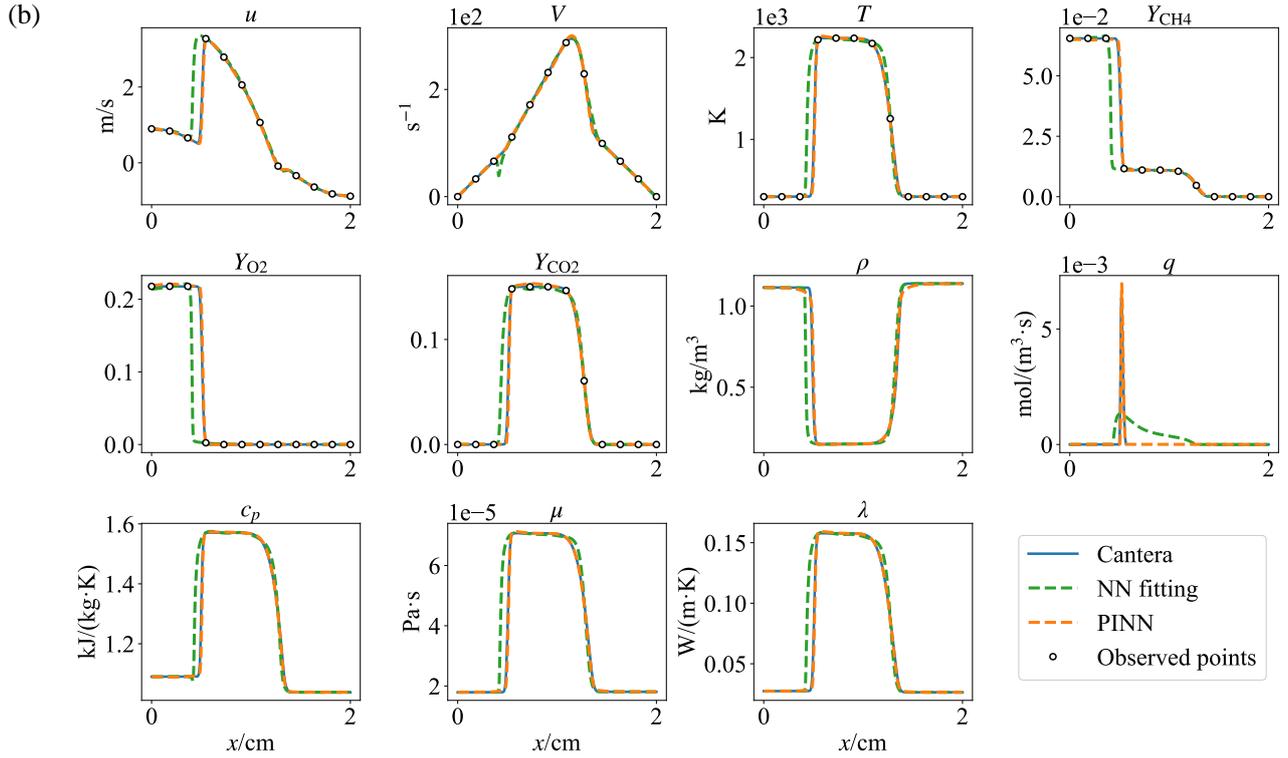

**Fig. 12.** Case 3 (inverse problem): Flame fields reconstructed by data fitting and PINN. The observation noise level is 1%. The data fitting is conducted with the same NN as PINN while the ODE losses are absent. (a) $\phi = 0.8$. $N_{ob} = 10$. (b) $\phi = 1.2$. $N_{ob} = 12$.

For the parameter inference, Table 7 shows that the AREs of inferred $E_a$ are all less than 2% with clean and 1% noisy observations, and less than 3% with 2% noisy observations. Note that the ARE of the initial value ($10^5$ J/mol) is 19.5%. These satisfactory results under various conditions illustrate the ability of PINN to infer chemical kinetics parameters from noisy sparse observations and with detailed combustion models. However, for $\Lambda$, the REs are all close to 14% under various $\phi$ and noise levels. The learning history curves of $\Lambda$ are also provided in Fig. 13, showing that its learning converges quickly under different $\phi$ and with different noise levels, although to values different from the reference values, which is not the case for $E_a$. We attribute this error to the intrinsic discretization error of the FDM adopted by Cantera. Note that the AD adopted by PINNs has no discretization error. Since the inference of $\Lambda$ relies entirely on the solving of the radial momentum equation, the residuals of this equation should be compared between PINN and Cantera. For PINN, the mean value of this residual (normalized), which is computed by AD, can all fall to $\mathcal{O}(10^{-3})$ under various $\phi$, demonstrating the accurate solution of this equation. However, for Cantera, it is impossible to compute the residuals by AD on discrete grids, and it doesn't make sense to compute the residuals by FDM. Therefore, all of the above factors (the stable "14%", the fast convergence, the accuracy of AD, and the small residual of PINN) point to a possible error in Cantera's solution of this equation, which is not our focus and needs to be further validated in the future.

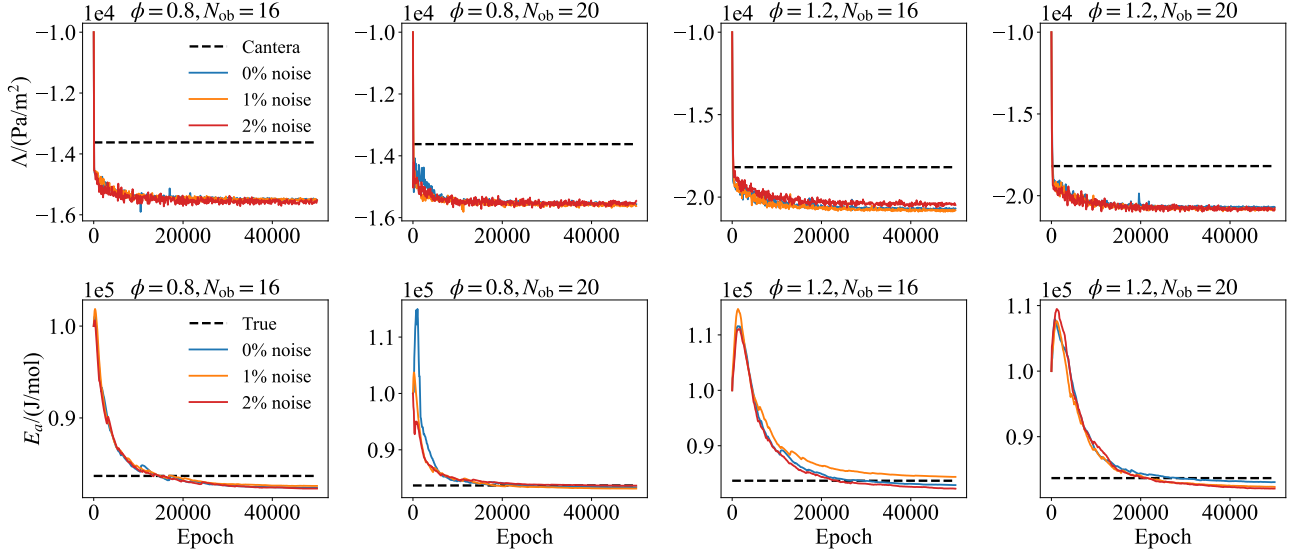

**Fig. 13.** Case 3 (inverse problem): Learning histories of pressure curvatures ($\Lambda$) and activation energies ($E_a$) under different $\phi$, $N_{ob}$, and noise levels.

Finally, the influence of observation sparsity on the PINN performance was investigated, as shown in Table 8. In general, the results are better for large $N_{ob}$ than for small $N_{ob}$. With increasing $N_{ob}$, the errors almost converge for $N_{ob} > 25$. For $N_{ob} < 16$, the results are not that bad, which is consistent with Fig. 12. As a reference, since the nozzle distance is 2 cm, $N_{ob} = 11$ means that the observation resolution is only 0.2 cm, which is quite larger than practical measurements [77]. Therefore, the significant few-shot learning ability of PINN with detailed combustion models is validated.

**Table 8**

Case 3 (inverse problem): $L_2$RE of PINN-reconstructed flame fields and RE of PINN-inferred activation energies ($E_a$) under different number of observed points ($N_{ob}$). The noise level is 0% and $\phi = 1.0$.

| $N_{ob}$ | $u$ | $V$ | $T$ | $Y_{CH4}$ | $\Lambda$ | $E_a$ |
|---|---|---|---|---|---|---|
| 10 | 6.91% | 0.45% | 4.52% | 5.82% | 14.10% | 0.47% |
| 12 | 3.43% | 0.54% | 2.87% | 3.38% | 13.89% | −1.60% |
| 14 | 4.48% | 0.62% | 3.09% | 3.99% | 14.31% | −0.56% |
| 16 | 4.37% | 0.34% | 3.68% | 4.25% | 13.63% | −1.17% |
| 18 | 2.49% | 0.17% | 1.72% | 2.39% | 14.04% | −1.26% |
| 20 | 5.26% | 0.29% | 3.59% | 4.54% | 13.89% | −0.42% |
| 25 | 2.10% | 0.15% | 1.44% | 2.09% | 14.00% | −1.27% |
| 30 | 1.22% | 0.09% | 1.80% | 2.23% | 13.33% | −0.05% |
| 40 | 0.91% | 0.08% | 1.05% | 1.47% | 13.72% | −0.51% |
| 50 | 1.04% | 0.11% | 0.81% | 1.33% | 13.64% | −0.37% |

### 3.3 Hyperparameter validation

In this subsection, we validate the effectiveness of the adopted hard constraints, normalization strategies, and NN architecture. Table 9 shows the results under different hard constraints, showing that without the hard constraint of $Y_k$, the results are worse and the species conservation is not exactly satisfied. For the hard constraint of $T$, it affects little and much when the hard constraint of $Y_k$ is adopted or not adopted, respectively. Nevertheless, the hard constraint of $T$ is strongly recommended. Otherwise, errors may occur during training, since an unphysical $T$ is obtained, which is often the situation of Case 1. Table 10 gives the results with different normalization strategies, showing that the traditional global normalization of $x$, which transforms the domain to [0, 1], yields completely wrong solutions. Only with the local normalization of the flame thin layers can PINN obtain satisfactory results [46]. The scaling of the dependent variables (**u**) has little impact on the results in Table 10, since they all follow the principle of magnitude normalization. Table 11 lists the metrics of PINN with different NN architectures, illustrating the effectiveness of MMLP in obtaining accurate solutions. The regular MLP fails to reduce the losses, thus yielding incorrect solutions. RWF has little impact on the results, possibly because that its effectiveness is more pronounced in higher-dimensional problems.

**Table 9**

Case 2 (forward problem): $L_2$RE of PINN-solved flame fields and RE of PINN-inferred laminar flame speeds ($s_L$) with different hard constraints. $R_{mean}$ and $R_{max}$ are the mean and maximum values of $|\sum_k Y_k - 1|$, respectively. No hard constraint of $T$ means not using the Sigmoid function, thus $T^\# = T^* = \alpha_T T$. No hard constraint of $Y_k$ means not using the Softmax function, thus $Y_k^\# = Y_k^* = \alpha_{Y_k} Y_k$. The $p = $ 5atm and $\phi = 1.6$.

| Hard constraint of | | $T$ | $Y_{CH4}$ | $u$ | $s_L$ | $R_{mean}$ | $R_{max}$ |
|---|---|---|---|---|---|---|---|
| $T$ | $Y_k$ | | | | | | |
| Yes | Yes | 0.31% | 0.20% | 0.70% | −0.49% | 3.71×10$^{-17}$ | 3.33×10$^{-16}$ |
| Yes | No | 2.80% | 3.64% | 8.52% | −5.89% | 2.35×10$^{-3}$ | 6.00×10$^{-3}$ |
| No | Yes | 0.32% | 0.32% | 0.57% | −0.91% | 3.84×10$^{-17}$ | 3.33×10$^{-16}$ |
| No | No | 33.45% | 19.54% | 93.88% | −92.04% | 3.46×10$^{-2}$ | 7.41×10$^{-2}$ |

**Table 10**

Case 2 (forward problem): $L_2$RE of PINN-solved flame fields and RE of PINN-inferred laminar flame speeds ($s_L$) with different normalization strategies (by changing $\alpha_x$, $\Delta_x$, and $\alpha_\mathbf{u}$). The $p = $ 5atm and $\phi = 1.6$.

| $x^*$ range | max($\mathbf{u}^*$) | $T$ | $Y_{CH4}$ | $u$ | $s_L$ |
|---|---|---|---|---|---|
| [−2, 2] | 5 | 0.31% | 0.20% | 0.70% | −0.49% |
| [−2, 2] | 1 | 0.31% | 0.26% | 1.16% | −0.36% |
| [0, 1] | 5 | 65.92% | 68.23% | 78.17% | 25.89% |
| [0, 1] | 1 | 67.97% | 71.48% | 78.33% | 21.73% |

**Table 11**

Case 2 (forward problem): $L_2$RE of PINN-solved flame fields and RE of PINN-inferred laminar flame speeds ($s_L$) with different NN architectures. The $p = $ 5atm and $\phi = 1.2$.

| NN | RWF | $T$ | $Y_{CH4}$ | $u$ | $s_L$ |
|---|---|---|---|---|---|
| MMLP | Yes | 0.25% | 0.43% | 0.51% | −0.53% |
| MMLP | No | 0.26% | 0.48% | 0.28% | −0.40% |
| MLP | Yes | 65.98% | 93.23% | 73.84% | 27.47% |
| MLP | No | 65.76% | 93.51% | 73.50% | 25.24% |

*3.4 Computational efficiency discussion*

Training of Case 1, Case 2, and Case 3 takes about 2.5 min, 75 min, and 140 min, respectively, on an NVIDIA GeForce RTX 3060 GPU. Their total training epochs are 30000, 30000, and 50000, respectively, so their training time per 100 epochs is 0.5 s, 15 s, and 16.8 s, respectively. As a comparison, for the FPP flames of Case 2, the required total epochs and training time per 100 epochs in a previous similar study [57] are 400000 and 19.8 s, respectively. For the reference solutions using traditional methods, the three cases averagely cost 1.5 s, 0.8 s, and 3.5 s on the Intel Core(TM) i7-12700F CPU, respectively, which is much faster than PINNs. This is consistent with the consensus that current PINNs have little efficiency advantage over traditional methods for forward problems [40, 78]. However, the mesh-free nature and zero discretization error remain the inherent strengths of PINNs. More importantly, PINNs still have two major advantages over traditional methods: Inverse problem solving and surrogate modeling. Firstly, the aforementioned time costs of traditional methods are all for forward problems. For inverse problems, such as field reconstruction and parameter inference, it is difficult for traditional methods to solve but straightforward for PINNs [63], so the above efficiency comparison is unfair for inverse problems. Secondly, PINNs have been shown to be more efficient for surrogate modeling of combustion systems [57]. To obtain solutions under different conditions, traditional methods have to compute them one by one, while PINNs can compute all cases simultaneously by setting the working conditions as NN inputs. After training, a continuous and differentiable surrogate model is obtained. In addition, for forward problems, although the proposed FlamePINN-1D framework is less efficient than traditional methods, which limits its application, it still has high research value because the essential findings may shed light on its application to inverse problems and surrogate modeling. A typical example can be Case 2 in this work, where the validated hyperparameters can be directly used in the inverse problems.

## 4. Conclusions

In this work, the FlamePINN-1D framework has been established to solve forward and inverse problems of 1D laminar flames based on PINNs. The governing equations of 1D laminar flames are integrated into the framework, playing a physical supervisory role in the learning process of PINNs, and enabling PINNs to solve both the forward and inverse problems in a unified manner. Three types of 1D laminar flames have been studied, including the freely-propagating and counterflow premixed flames. The main conclusions are:

(1) For forward problems, FlamePINN-1D can accurately solve the flame fields and simultaneously infer the laminar flame speeds of freely-propagating premixed flames under various working conditions. Compared to traditional numerical methods, FlamePINN-1D is differentiable and mesh-free, and has no discretization errors. These advantages can, to some extent, compensate for its shortcomings in computational efficiency for forward problems.

(2) For inverse problems, FlamePINN-1D can accurately reconstruct the continuous fields and infer the transport properties and chemical kinetics parameters from noisy sparse observations, which opens the gate to optimizing chemical mechanisms from observations of laboratory 1D flames. The field reconstruction is more accurate than pure data-fitting, demonstrating the effectiveness of the physical constraints in FlamePINN-1D.

(3) The proposed warmup pretraining strategy, hard constraints, thin-layer normalization, and modified NN architectures have been proven to be essential for the robust learning of FlamePINN-1D, which can serve as fundamental experiences for the future development of PINNs in solving combustion problems.

There are still some shortcomings in the current framework, such as the robustness, theoretical analysis, and applicability to more practical flames. To address these limitations and further explore the ability of FlamePINN-1D, some possible future work includes: (1) Adopting more detailed chemical mechanisms to improve the complexity and practicality of the problems. (2) Applying FlamePINN-1D to more types of flames, such as diffusion flames, to broaden its applicability. (3) Inferring more unknown parameters and quantifying the inference uncertainty to enhance model reliability. (4) Exploring the feasibility of FlamePINN-1D to study flame limits, stretch, and flamelet tabulation, etc.

**Declaration of competing interest**

The authors declare that they have no known competing financial interests or personal relationships that could have appeared to influence the work reported in this paper.

**Acknowledgments**

This work is supported by the National Key R&D Program of China (NO. 2021YFA0716202) and the National Natural Science Foundation of China (Grant No. 62273149).

## Appendix A. Algorithm for the reference solution of Case 1

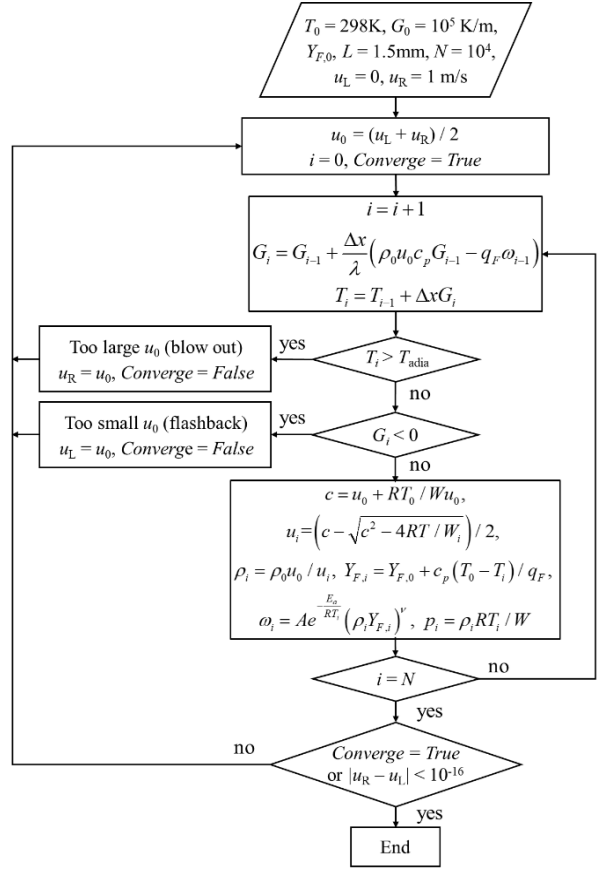

**Fig. 14.** The algorithm, which is based on the FDM and bisection method, for obtaining the reference solution of Case 1. The inlet velocity ($u_0$) is also the laminar flame speed ($s_L$).